\documentclass[onecolumn,journal]{IEEEtran}
\usepackage{amsmath,amsfonts}
\usepackage{algorithmic}
\usepackage{array}
\usepackage[caption=false,font=normalsize,labelfont=sf,textfont=sf]{subfig}
\usepackage{textcomp}
\usepackage{stfloats}
\usepackage{url}
\usepackage{verbatim}
\usepackage{graphicx}
\usepackage{cite}
\usepackage{newtxtext,newtxmath} % 设置 Times New Roman 字体

\usepackage{float}  % <-- 加上这个
\usepackage{tikz}
\usetikzlibrary{fit}
\usepackage{float}
\usepackage{placeins} 
\usepackage{color}
\usepackage{cleveref}
\usepackage{multirow}
\usepackage{longtable}

\usepackage{booktabs}
\usepackage{setspace}
\renewcommand{\figurename}{Fig.}

\ifCLASSINFOpdf
\else
\fi
\begin{document}
\title{SFFR: Spatial-Frequency Feature Reconstruction for Multispectral Aerial Object Detection}
\author{Xin~Zuo,
        Chenyu~Qu,
       Haibo~Zhan,
       Jifeng~Shen and Wankou~Yang
       % <-this % stops a space
%}

\thanks{X. Zuo and C. Qu are with the School of Computer, Jiangsu University of Science and Technology, Zhenjiang, 212003 China.}% <-this % stops a space
\thanks{H. Zhan and J. Shen are with the School of Electrical and Information Engineering, Jiangsu University, Zhenjiang, 212013 China (e-mail: shenjifeng@ujs.edu.cn).}% <-this % stops a space
\thanks{W. Yang is with the School of Automation, Southeast University, Nanjing, 210096 China.}% <-this % stops a space
}

\markboth{Journal of \LaTeX\ Class Files,~Vol.~13, No.~9, September~2025}%
{Shell \MakeLowercase{\textit{et al.}}: SFFR: Spatial-Frequency Feature Reconstruction for Multispectral Aerial Object Detection}

\maketitle

\begin{abstract}
Recent multispectral object detection methods have primarily focused on spatial-domain feature fusion based on CNNs or Transformers, while the potential of frequency-domain feature remains underexplored. 
In this work, we propose a novel Spatial and Frequency Feature Reconstruction method (SFFR) method, which leverages the spatial-frequency feature representation mechanisms of the Kolmogorov–Arnold Network (KAN) to reconstruct complementary representations in both spatial and frequency domains prior to feature fusion.
The core components of SFFR are the proposed Frequency Component Exchange KAN (FCEKAN) module and Multi-Scale Gaussian KAN (MSGKAN) module. 
The FCEKAN introduces an innovative selective frequency component exchange strategy that effectively enhances the complementarity and consistency of cross-modal features based on the frequency feature of RGB and IR images.
The MSGKAN module demonstrates excellent nonlinear feature modeling capability in the spatial domain. By leveraging multi-scale Gaussian basis functions, it effectively captures the feature variations caused by scale changes at different UAV flight altitudes, significantly enhancing the model’s adaptability and robustness to scale variations.
It is experimentally validated that our proposed FCEKAN and MSGKAN modules are complementary and can effectively capture the frequency and spatial semantic features respectively for better feature fusion.
Extensive experiments on the SeaDroneSee, DroneVehicle and DVTOD datasets demonstrate the superior performance and significant advantages of the proposed method in UAV multispectral  object perception task. Code will be available at https://github.com/qchenyu1027/SFFR.
\end{abstract}


% \fi
\IEEEpeerreviewmaketitle

\section{Introduction}
\IEEEPARstart{I}{n} recent years, the rapid development of unmanned aerial vehicle (UAV) technology has significantly improved the efficiency and flexibility of aerial image acquisition, providing abundant and high-quality data resources for object detection tasks in computer vision. Aerial object detection tasks~\cite{wang_improved_2023, fu_point-based_2021,han_redet_2021} have shown broad application prospects in various real-world scenarios, such as intelligent traffic monitoring~\cite{sun_rsod_2022}, environmental surveillance~\cite{ramachandran_review_2021}, and military reconnaissance~\cite{liu_military_2022}, attracting extensive research attention. However, compared to ground-level images, aerial images typically exhibit more complex background, significant scale variations, and sparsely distributed objects, all of which pose substantial challenges to the robustness and discriminative capacity of existing detection models. Traditional single-modal detection frameworks often suffer from feature degradation and performance drops under complex conditions such as low illumination, occlusion, or dense object layouts.

\begin{figure}[H]
\centering
\captionsetup{justification=justified} 
\includegraphics[width=1\textwidth]{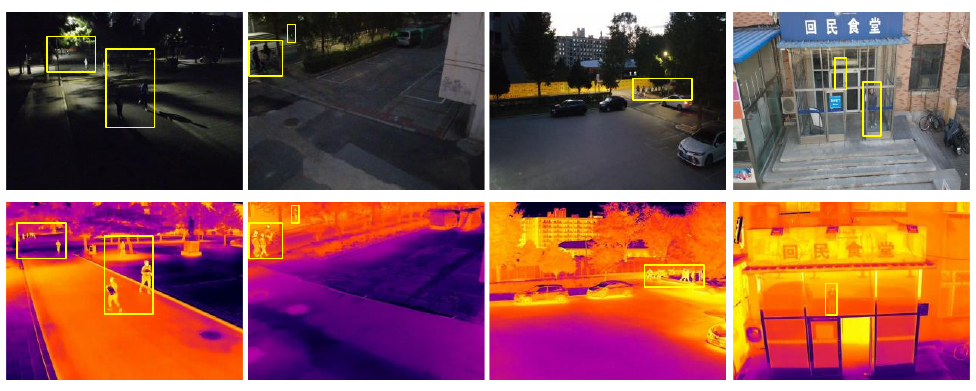}

\caption{ Examples of RGB–IR image pairs from the DVTOD dataset. The first and second rows show RGB and IR images respectively. RGB images excel at capturing color and texture information through visible light reflection, whereas IR images specialize in visualizing heat radiation. 
The two modalities complement each other, especially in challenging conditions such as low illumination, occlusion, or complex backgrounds, thereby enhancing object visibility and discriminability.}
\label{fig1}
\end{figure}

To address the challenges in UAV-based object detection, multispectral imaging and cross-modal as well as multi-scale feature fusion have become key research focuses. As illustrated in Fig.~\ref{fig1}, RGB and IR modalities exhibit inherent complementarity in visual perception: the RGB modality captures rich texture and color details under favorable lighting conditions, whereas the IR modality demonstrates stronger target sensitivity in low-light or occluded environments. However, in practical UAV applications, varying flight altitudes and viewing angles induce significant multi-scale variations in target size and structural features, substantially increasing detection difficulty. Effectively integrating these heterogeneous modalities not only enhances feature representation but also improves the robustness of cross-modal learning. Although existing cross-modal feature fusion methods~\cite{ren_faster_2017, yang_cascaded_2023, peng_hafnet_2023, lee_insanet_2024, xing_ms-detr_2024} have exhibited promising progress, they still struggle to fully exploit the synergistic potential of modality complementarity and scale adaptation in the presence of large scale and structural variations as well as complex background interference, resulting in performance degradation.

To address these limitations, it is essential to take a closer look at the current fusion approaches, which can generally be categorized into two main paradigms. The first category encompasses CNN-based approaches~\cite{ding_convolutional_2020, jung_fusionnet_2020},  which demonstrate robust local feature modeling capabilities but are inherently constrained by limited receptive fields.
The second category involves Transformer-based methods~\cite{chen_transformer_2024, qingyun_cross-modality_2022}, which excel at capturing long-range dependencies but typically require deep stacking to enhance representation capabilities and suffer from high computational complexity. 
Moreover, existing approaches focus primarily on spatial-domain feature modeling and tend to overlook the crucial role of different frequency components in achieving effective multi-scale and cross-modal fusion.
Therefore, effectively integrating frequency–spatial domain feature modeling capabilities to design more efficient, robust, and expressive cross-modal and multi-scale feature fusion schemes remains a critical and open research challenge.

In this paper, a novel Spatial-Frequency Feature Reconstruction method, dubbed SFFR, which explicitly reconstructs complementary representations in both spatial and frequency domains prior to feature fusion. SFFR is built upon the feature representation mechanisms of the Kolmogorov–Arnold Network (KAN). Unlike traditional approaches, KAN is inherently capable of capturing high-order nonlinear interactions, particularly in the frequency domain. Its flexible formulation based on learnable univariate functions enables precise alignment of structural differences between IR and RGB modalities across both spectral and spatial domains. At the core of this method is the newly designed Frequency Component Exchange KAN (FCEKAN) module. Building upon the Fourier-based KAN formulation, FCEKAN exhibits powerful frequency-domain approximation and nonlinear modeling capabilities. Specifically, RGB and IR features are first projected into the frequency domain to enable frequency-aware representation learning. A selective frequency component exchange strategy is then employed to extract high-frequency edge and contour features from the IR modality while preserving rich texture and structural details from the RGB modality. This mechanism enhances the complementarity and consistency between modalities, significantly improving cross-modal fusion performance.
To further enhance spatial representation capabilities, We propose the MSGKAN module to enhance spatial representation by combining the nonlinear power of multi-scale Gaussian radial basis functions with the local modeling ability of CNNs. This module effectively captures multi-scale spatial features, complementing the frequency-domain pathway. Additionally, MSGKAN robustly models feature variations caused by scale changes at different UAV flight altitudes, further improving the model’s adaptability to scale variations. Working together, our FCEKAN and MSGKAN modules greatly enhance the multi-modal feature representation of SFFR for multispectral aerial object detection. Extensive experiments on the SeaDroneSee~\cite{varga_seadronessee_2021}, DroneVehicle~\cite{sun_drone-based_2021}, and DVTOD~\cite{song_misaligned_2024} demonstrate the effectiveness of the proposed method. 

In summary, our main contributions are as follows:
\begin{enumerate}

    \item 
    A novel spatial–frequency feature reconstruction (SFFR) method that integrates spatial and frequency domain representations, and demonstrate its effectiveness on the task of aerial object detection by enhancing the complementarity between RGB and IR features.
    
    \item A frequency component exchange mechanism (FCEKAN) is proposed to fully harness modality-specific information by effectively integrating complementary spectral features across modalities in the frequency domain. 
  
    \item By fully leveraging the nonlinear modeling capability of multi-scale Gaussian radial basis functions within the MSGKAN module, our model effectively captures complex spatial features and adapts to scale variations across UAV flight altitudes.
    
    \item We conduct experiments on SeaDroneSee, DroneVehicle and DVTOD datasets. Experimental results show our SFFR consistently achieves state-of-the-art results.
\end{enumerate}

\par The rest of this paper is organized as follows. \Cref{sec2} reviews related work on multispectral and aerial object detection.  \Cref{sec3} describes the proposed method. \Cref{sec4} presents the experimental results and analysis. Finally, we summarize the paper in \Cref{sec5}.

\section{Related Work}
\label{sec2}
\subsection{Multispectral Object Detection}

Recent years have witnessed continuous advancements in multispectral object detection~\cite{shen_icafusion_2024, liu_aerial_2025, li_specdetr_2025, fu_cf-deformable_nodate,  jang_multispectral_2025, xu_enhanced_2024, yang_multidimensional_2025, zheng_mcafnet_2025, peng_coxnet_2025, jang_camdet_2025}. Among these methods, ICAFusion~\cite{shen_icafusion_2024} proposes an iterative cross-attention fusion framework for multispectral object detection. By incorporating dual cross-attention modules and iterative feature refinement, it efficiently captures the complementary information between RGB and IR modalities. ADMPF~\cite{liu_aerial_2025} proposes an RGB–IR multibranch progressive fusion method for aerial image object detection, aiming to fully exploit the complementary information to improve detection accuracy. In parallel, Transformer-based approaches have also been explored. SpecDETR~\cite{li_specdetr_2025} presents a Transformer-based framework for the detection of hyperspectral point object, introducing a self-excited subpixel-scale attention mechanism along with a hybrid label assigner. CF-Deformable DETR~\cite{fu_cf-deformable_nodate} proposes an alignment-free architecture based on cross-modal deformable attention and loss of point-level feature consistency for RGB-IR object detection. Beyond these, Jang et al.~\cite{jang_multispectral_2025} introduce the MCOR module, which integrates CIC and CSCR to enhance modality complementarity and emphasize object-centric features, achieving strong performance on multiple multispectral benchmarks. Moreover, frequency- and spatial-aware methods have been proposed. Xu et al.~\cite{xu_enhanced_2024} propose an Enhanced Spectral--Spatial Fusion Network , which introduces a Spectral-Spatial Enhancement (SSE) module and a fast Fourier transform (FFT)-based fusion module to improve cross-modal feature consistency and enhance small object detection performance.  Moreover, multidimensional and adaptive fusion strategies have been explored. Yang et al.~\cite{yang_multidimensional_2025} propose the Multidimensional Fusion Network (MMFN), which fuses RGB and IR features from local, global and channel perspectives to generate highly discriminative cross-modal representations. MCAFNet~\cite{zheng_mcafnet_2025} adaptively fuses RGB, IR and concatenated features via CMIT, MAWF and 3D-IAFE modules to enhance local and global perception, achieving state-of-the-art results on multiple datasets. Beyond multispectral object detection, RGB-Infrared (RGB-I) salient object detection (SOD) have seen extensive study~\cite{wang_cross-modal_2024, wang_mirror_2025, xie_cross-modality_2023, gao_highly_2025} because they likewise exploit cross-modal fusion to highlight informative regions. CAINet~\cite{wang_cross-modal_2024}, which adopts a dual-branch Swin Transformer with CIEM for co-attention, MAID for multi-scale fusion and EEM for edge enhancement. Although SOD emphasizes saliency map generation rather than bounding-box localization and class prediction, it shares core challenges with multispectral detection-most notably cross-modal feature alignment and multi-scale representation learning. Motivated by these studies, which emphasize cross-modal complementarity and multi-scale representation learning, we propose a novel KAN-based cross-modal and multi-scale fusion framework. Our approach explicitly reconstructs complementary spatial–frequency representations, thereby enhancing feature alignment between RGB and IR modalities and improving multispectral object detection performance.

\subsection{Aerial Object Detection}
Aerial object detection techniques have evolved through both traditional computer vision and deep learning paradigms. In recent years, several aerial object detection methods based solely on RGB modalities have achieved promising results~\cite{ding_learning_2019, xu_gliding_2021, lir32019, kim_ecap-yolo_2021,  peng_feature-enhanced_2024}. Ding et al.~\cite{ding_learning_2019} propose the RoI Transformer module, which converts HRoIs to RRoIs through spatial transformation, thus improving the accuracy of object detection. R\textsuperscript{3}-Net~\cite{lir32019} addresses the detection of arbitrarily oriented vehicles in aerial imagery and videos through the generation of rotatable rectangular bounding boxes. Kim et al.~\cite{kim_ecap-yolo_2021} propose a simple and effective framework for multi-directional object detection, such as objects and scene text in aerial images. To further enhance detection performance, several approaches have explored the integration of multispectral data—particularly the fusion of RGB and IR modalities~\cite{qingyun_cross-modality_2022, hu_improving_2023, sun_drone-based_2022}. C\textsuperscript{2}former~\cite{yuan_mathbfc2former_2024} proposes a novel calibrated and complementary Transformer, that is used for RGB-IR object detection to solve the problems of inaccurate modal calibration and inaccurate fusion. In this work, we build an aerial object detection framework incorporating our proposed Cross-modal and multi-scale fusion module, aiming to improve detection accuracy and robustness under complex aerial imaging conditions.

\subsection{Kolmogorov-Arnold Network}
With the increasing integration of advanced mathematical tools in computer vision, many complex modeling challenges can now be addressed more effectively. Kolmogorov–Arnold Networks~\cite{liu_kan_2025} proposes an approach that differs from the traditional Multi-layer Perceptron (MLP) approach of using fixed activation functions at the nodes by introducing learnable activation functions at the edges of the network. U-KAN~\cite{li_u-kan_2024} extends this concept by integrating KAN into the U-Net architecture, replacing conventional convolutional blocks with the Tok-KAN module and demonstrating improved performance in image segmentation and generation tasks. GraphKAN~\cite{zhang_graphkan_2024} replaces the MLP with KAN to demonstrate its advantages in feature extraction and further improves node classification accuracy by substituting LayerNorm with a KAN-compatible alternative. Kanremote~\cite{cheon_kolmogorov-arnold_2024} combines KAN with various pretrained convolutional backbones, demonstrating strong performance in remote sensing scene classification through parameterized configurations. KAT~\cite{yang_kolmogorov-arnold_2024} introduces rational basis functions to replace the B-spline functions originally used in KAN. Compared with B-splines, which are less efficient on GPUs, rational functions are expressed as ratios of polynomials, enabling more efficient computation and stronger approximation capability, thereby achieving competitive performance in large-scale vision tasks such as image recognition and semantic segmentation. Compared with conventional CNNs and Transformers, which are constrained by fixed convolutional kernels or attention mechanisms, KAN offers greater flexibility by employing learnable basis functions that can approximate complex nonlinear mappings. This property allows KAN to jointly model spatial- and frequency-domain interactions within a unified framework, making it particularly advantageous for multispectral object detection tasks that demand fine-grained spatial–frequency fusion. In this work, we incorporate KAN into a multispectral object detection framework to enhance nonlinear semantic fusion and promote deeper interaction between RGB and IR modalities.

\section{Proposed Method}
\label{sec3}
\subsection{Preliminary}
The core idea of KAN is to decompose complex multivariate functions into simpler univariate components, allowing for effective function approximation.
Specifically, it is based on the Kolmogorov-Arnold representation theorem, which states that a multivariate continuous function $f(x_1,x_2,\cdots,x_n)$ can be expressed as a combination of a series of univariate functions, formulated mathematically as:
\begin{equation}
f(x_1,x_2,\cdots,x_n) = \sum_{q=1}^{2n+1} \Phi_q\left( \sum_{p=1}^{n} \phi_{q,p}(x_p) \right)
\label{eq1}
\end{equation}
where $\phi_{q,p}:[0,1] \rightarrow \mathbb{R}$ is a nonlinear transformation of the input variable $x_p$, mapping the \(p\)-th input feature to the \(q\)-th output feature, similar to the activation function in a neural network. The $\Phi_q:\mathbb{R} \rightarrow \mathbb{R}$ combines the summation of the inner functions. Each KAN layer is defined by a matrix of learnable one-dimensional functions:
\begin{equation}
\Phi = \{\phi_{q,p}\}
\end{equation}
where \(p = 1, \ldots, n\) and \(q = 1, \ldots, 2n+1\). These functions are jointly learned to realize a nonlinear mapping between input and output features.

As shown in Eq.~\ref{eq6}, $\phi(x)$ is the activation function for each edge, which mainly consists of a weighted linear combination of the B-spline and the basis function:
\begin{align}
\phi(x)   &= w_1\, \phi_1(x) + w_2\, \phi_2(x) \label{eq6} \\
\phi_1(x) &= \mathrm{Silu}(x) = \frac{x}{1 + e^{-x}} \label{eq7} \\
\phi_2(x) &= \mathrm{Spline}(x) = \sum_i c_i B_i(x) \label{eq8}
\end{align}
where \( w_1 \) and \( w_2 \) are used to control the overall size of the activation function.  \( \mathrm{Spline}(\cdot) \)  can be parameterised as a combination of the basis functions of B-splines. \( c_i \) is trainable and \( B_i(x) \) is a predefined B-splines basis function.

The entire KAN is implemented by function compositing. For a KAN containing $L$ layers with an input vector $x$, the final output is:\\[-0.5em]
\begin{equation}
KAN(x) = (\Phi_{L-1} \circ \Phi_{L-2} \circ \cdots \circ \Phi_0)(x)
\end{equation}
where \( \Phi_L \) is the function matrix of the \( L \)-th layer, and \( \circ \) denotes the function composite.

\subsection{Architecture}

\begin{figure}[t]
\centering
\captionsetup{justification=justified} 
\includegraphics[width=0.9\textwidth]{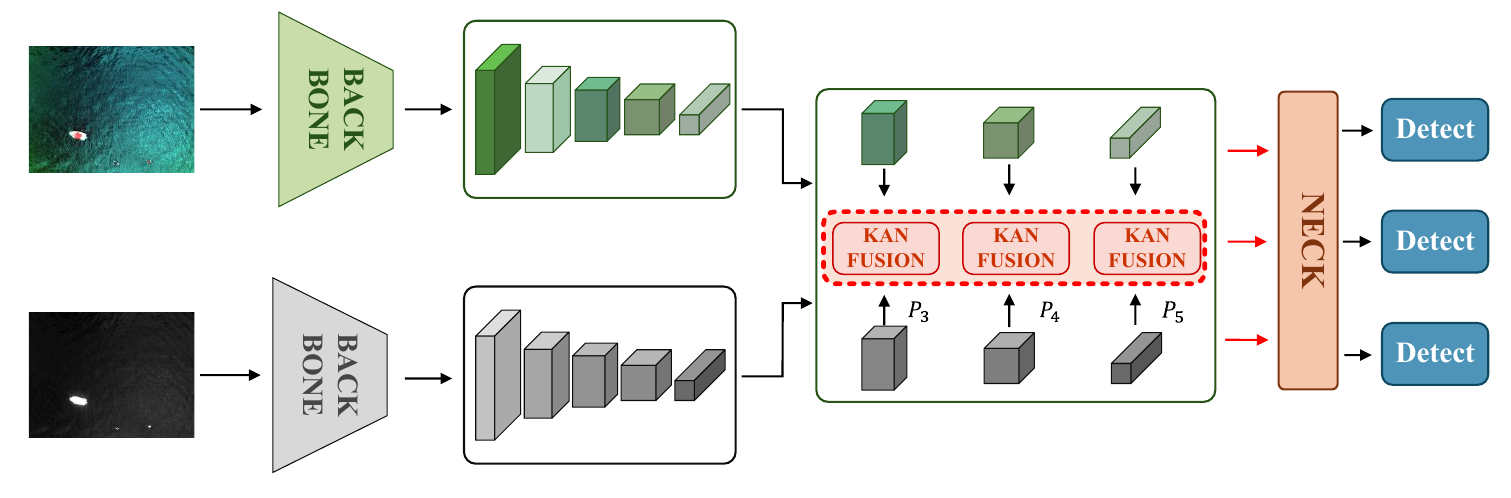}
\caption{Overview of the proposed multispectral aerial object detection framework. (The upper and lower branches extract RGB and IR features, respectively. Cross-modal and multi-scale fusion is performed via the proposed KANFUSION module. Subsequently, a multi-scale fusion network in the Neck aggregates and enhances the fused features, and the Detection Head produces the final detection results.)}
\label{fig.2}
\end{figure}

The architecture of the proposed multispectral aerial object detection framework is illustrated in Fig.~\ref{fig.2}, comprising three principal components: Backbone, Neck, and Detection Head. The Backbone extracts fundamental feature representations from input RGB and IR images. To leverage complementary information across modalities, cross-modal and multi-scale feature fusion (KANFusion) is performed at intermediate network stages. 

\begin{figure}[t]
\centering
\captionsetup{justification=justified} 
\includegraphics[width=0.9\textwidth]{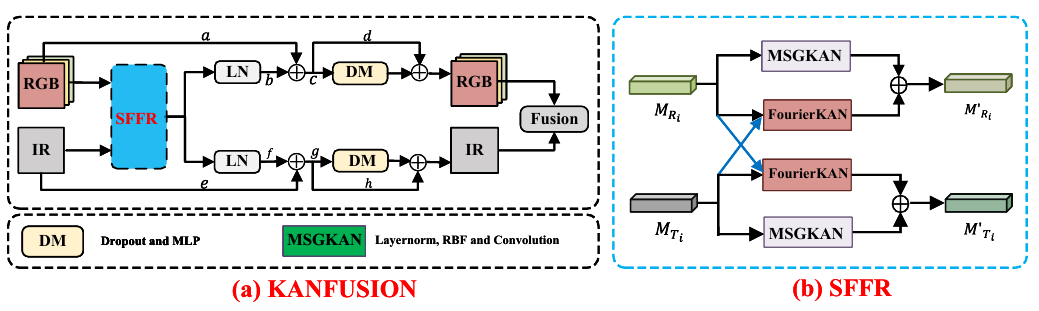}
\caption{Illustration of our cross-modal feature fusion module, which consists of two parallel feature extraction networks. The overall fusion module KANFUSION (a) integrates RGB and IR features via the SFFR (b) module, whose detailed architecture is shown on the right.}
\label{fig.3}
\end{figure}

As illustrated in Fig.~\ref{fig.3}, KANFusion (a) first employs the SFFR (b) module, producing enhanced representations. The detailed structure of the SFFR module is described in a subsequent Section~\ref{sec:SFFR}. Subsequently, the original features are fused with the Layernorm enhanced features via learnable coefficients \(a, b, e, f\), and each is further mapped through a Dropout-equipped MLP to introduce nonlinearity. Finally, the mapped features are combined with residual features weighted by learnable coefficients \(c, d, g, h\) to complete the feature update.

To generate the final detection results, the Neck and Detection Head progressively enhance and integrate features from earlier stages. Commonly used backbone networks, such as CSPDarknet53~\cite{bochkovskiy_yolov4_2020}, ResNet~\cite{he_deep_2015}, and VGG16~\cite{simonyan_very_2015}, offer strong feature extraction capabilities. In our design, we employ a dual-stream architecture built upon YOLOv5, chosen for its maturity, stability, and broad adoption in object detection research. Both modality branches employ CSPDarknet53 as the backbone, which effectively balances accuracy and computational efficiency. The Neck incorporates FPN and PANet to facilitate multi-scale feature fusion, while the Detection Head adopts the YOLO to perform accurate and efficient object localization and classification.

\subsection{Intra-modality Feature Enhancement}
\subsubsection{Multiscale Nonlinear Feature Transform via MSGKAN}
\label{sec:MSGKAN feature enhancement}
Inspired by recent advances in nonlinear function modeling~\cite{drokin_kolmogorov-arnold_2024, cheon_kolmogorov-arnold_2024,  li_kolmogorov-arnold_2024}, the feature enhancement module is built upon the KAN framework and is specifically designed for intra-modal feature enhancement in the spatial domain, as illustrated in Fig.~\ref{five_structures}(a). First, LayerNorm is applied to normalize the input features, which helps stabilize the training process. Then, we map the normalized features to a high-dimensional nonlinear latent space, which is achieved through our innovative multi-scale Gaussian radial basis function. Specifically, we introduce Gaussian functions with different bandwidths to extract response features at multiple scales and aggregate them to form a unified representation. The responses are aggregated using a set of learnable weights \( w_{nj} \), enabling the model to adaptively emphasize discriminative components and enhance the representation of key semantic patterns. 

Subsequently, convolution operation is employed to extract local spatial features, effectively integrating nonlinear transformation with spatial structure modeling.
 Given the input feature \( M_i \in \mathbb{R}^{B \times H \times W \times C_{\text{in}}} \) and its corresponding output \( M'_i \in \mathbb{R}^{B \times H \times W \times C_{\text{out}}} \), $B$ represents the batch size, $H$ and $W$ denote the height and width of the feature map, respectively, while $C_{in}$ and $C_{out}$ refer to the number of input and output channels. The process can be formulated  as:\\[-1em]
\begin{align}
f_{\mathrm{gus}}(M_i) 
&= \sum_{n=1}^{N} \sum_{j=1}^{K} w_{nj} \, \phi_j\left(\left\| M_i - c_n \right\|\right)
\label{eq:MSGKAN_1} \\[-0.3em]
M'_i 
&= \text{Conv}\left(f_{\mathrm{gus}}(M_i)\right)
\label{eq:MSGKAN_2} \\[-0.3em]
\phi_j(r) 
&= \exp\left(-\frac{r^2}{2h_j^2}\right)
\label{eq:MSGKAN_3}
\end{align}
where $w_{nj}$ denotes the learnable weights obtained via the convolution layer, 
and $\phi_j(\cdot)$ denotes the multi-scale radial basis function (MSRBF). Each MSRBF is parameterized by a scale parameter $h_j$, which controls its receptive bandwidth. $K$ is the number of scale parameters $h_j$ used in the MSRBF, fixed by the predefined list of $h_j$. Therefore, we conduct ablation studies on the scale parameter $h$ (see Table~\ref{scale-param}), while keeping $K$ fixed to ensure stable and fair evaluation. \( \|M_i - c_n\| \) denotes the distance between the input point $M_i$ and the center point $c_n$, which is known as the radial distance, and \( N \) is the number of MSRBF centers. $\text {Conv}(\cdot)$ represents convolution operation, $ \exp(\cdot)$ represents the exponential function. r is the radial distance.

\subsubsection{Non-local Efficient Feature Transform via FourierKAN}
\label{sec:FourierKAN feature enhancement}
As shown in Fig.~\ref{five_structures}(b), we innovatively introduce FourierKAN~\cite{xu_fourierkan-gcf_2024} to enhance the features of RGB and IR modalities from a frequency-domain perspective. Specifically, unlike traditional spatial-domain operations, FourierKAN constructs nonlinear mappings by superimposing multiple Fourier basis functions, enabling a modeling strategy that is both expressive and compact. We first project the RGB and IR features into the frequency domain and then apply the Fourier transform to decompose the input features into a summation of Sine and Cosine basis functions that span all channels and frequency components. Frequency-domain processing transforms complex spatial patterns into interpretable frequency components, effectively modeling long-range dependencies such as repetitive textures and continuous edges. This process provides more stable and discriminative semantic feature representations for subsequent tasks.
Given an input feature map \( M_{i} \in \mathbb{R}^{B \times C \times H \times W} \), we first flatten the spatial dimensions and reshape it into a 2D matrix \( M_{i} \in \mathbb{R}^{N \times d} \), where \( N = B \times H \times W \) denotes the total number of spatial locations across the batch, and \( d = C \) represents the input feature dimension. The corresponding formulations are defined as:\\[-0.8em]
\begin{align}
f_{\mathrm{fourier}}(M_{i}) &= \sum_{j=1}^{d} \sum_{k=1}^{g} \big( 
\cos(k M_{ij}) \cdot a_{jk} \notag \\[-0.3em]
&\quad + \sin(k M_{ij}) \cdot b_{jk} \big) \label{eq:fourier_feature} \\[-0.3em] % �� 手动压缩行距
M'_i &= f_{\mathrm{fourier}}(M_{i}) \label{eq:fourier_output}
\end{align}
\noindent
where \( d \) denotes the input feature dimension, \( a_{jk} \in \mathbb{R}^{d \times g} \) and \( b_{jk} \in \mathbb{R}^{d \times g} \) are learnable parameter matrices. The functions \( \sin(\cdot) \) and \( \cos(\cdot) \) denote the sine and cosine basis functions, respectively. The parameter \( g \) determines the number of \( \sin(\cdot) \) and \( \cos(\cdot) \) terms in the Fourier series expansion~\cite{brigham_fast_1967}. In our implementation, the parameter $g$ is defined as equivalent to the gridsize, and its influence is systematically assessed through an ablation study conducted with different gridsize values, as reported in Table~\ref{gridsize}.

\subsection{Cross-modal Feature Enhancement}
\subsubsection{Fourier-based Cross Enhancement KAN (FCEKAN)}
\label{sec:FourierKAN Component Exchange}
Building on FourierKAN's strengths in handling both intra-modal features (section~\ref{sec:FourierKAN feature enhancement}), we further propose an innovative frequency component exchange mechanism, termed FCEKAN, to enhance cross‑modal feature fusion, as shown in Fig.~\ref{five_structures}(d). By transforming features from the spatial domain to the frequency domain, the model can more effectively distinguish low-frequency and high-frequency components in the RGB and IR modalities. In the frequency domain processing, we introduce a Fourier feature component interaction strategy, which involves swapping the mapping channels of Sine and Cosine basis functions within a specific frequency range to achieve cross-reconstruction of feature components. This design not only captures the structural and textural characteristics within each individual modality across different frequency scales, but also effectively establishes structural complementarity and feature correlations across modalities. The formulas are defined as follows:

\begin{align}
f_{\mathrm{Cross}}(M_{R_{i}},M_{T_{i}}) &= \sum_{j=1}^{d} \sum_{k=1}^{g} \left( 
\cos(k M_{R_{ij}}) \cdot a_{jk} \right. \notag \\[-0.3em]
&\quad \left. + \sin(k M_{T_{ij}}) \cdot b_{jk} \right) \label{eq:CrossF1} \\[-0.3em]
M'_{R_{i}} &= f_{\mathrm{Cross}}(M_{R_{i}}, M_{T_{i}}) 
\end{align}

\noindent
where \( M_{R_{i}} \) and \( M_{T_{i}} \) represent the input features. The variable \( d \) denotes the feature dimension, \( M_{R_{ij}} \) and \( M_{T_{ij}} \) refer to the \( j \)th feature component of \( M_{R_{i}} \) and \( M_{T_{i}} \), respectively. The matrices \( a_{jk} \in \mathbb{R}^{d \times g} \) and \( b_{jk} \in \mathbb{R}^{d \times g} \) are learnable parameters. The functions \( \sin(\cdot) \) and \( \cos(\cdot) \) denote the sine and cosine basis functions. The hyperparameter \( g \) controls the number of terms in the Fourier series expansion~\cite{brigham_fast_1967}, which is determined by the predefined frequency resolution (gridsize).

\subsubsection{Implicit Cross-modal And Multi-scale Feature Enhancement}
\begin{figure}[t]
    \centering
    \captionsetup{justification=justified} 
    \includegraphics[width=0.8\textwidth]{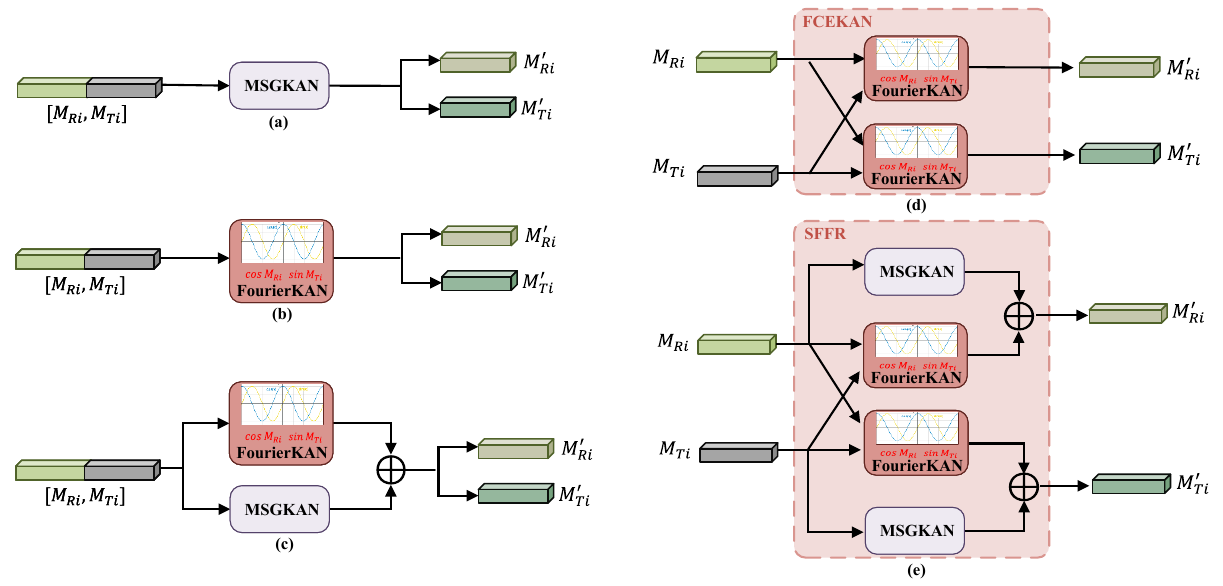} 
    \caption{Overview of the proposed five structures: from left to right are the Intra-modality feature enhancement module, the cross-modal feature enhancement module, the FCEKAN module, and finally the SFFR module, which constitutes our final method.} 
    \label{five_structures}
\end{figure}

We employ MSGKAN and FourierKAN to model the RGB and IR features separately, as illustrated in Fig.~\ref{five_structures}(c). MSGKAN leverages multi‑scale Gaussian-kernel‑based nonlinear mappings to capture fine‑grained, scale‑aware spatial characteristics from RGB features, while FourierKAN emphasizes the extraction of global structural information from IR features in the frequency domain. By exploiting the complementary nature of these spatial and spectral representations, this design facilitates more robust and discriminative cross‑modal feature enhancement.

\textbf{FourierKAN For IR Features:}
IR features primarily encode thermal and structural information and often lack fine-grained high‑frequency details, yet they carry rich global semantics. The FourierKAN module effectively separates low‑frequency structural components from high‑frequency noise in the frequency domain, thereby enhancing the feature representation of IR inputs. This approach is particularly advantageous for capturing thermal patterns and global shape information, yielding more accurate and robust feature representations.

\textbf{MSGKAN For RGB Features:}
We introduce MSGKAN, a multi‑scale, Gaussian‑kernel‑based nonlinear mapping module that operates on RGB inputs to capture fine‑grained, scale‑aware spatial characteristics. In MSGKAN, convolution layers encode local spatial information, while nonlinear mappings further refine complex semantic patterns across varying scales. This design strengthens the model’s ability to adapt to scale variations caused by UAV altitude changes, yielding richer and more discriminative RGB feature representations. Moreover, MSGKAN promotes structural complementarity with IR features, facilitating more effective cross‑modal feature fusion.

Given the input features \( M_{T_{i}} \in \mathbb{R}^{B \times H \times W \times C_{\mathrm{in}}} \) and \( M_{R_{i}} \in \mathbb{R}^{B \times H \times W \times C_{\mathrm{in}}} \), which denote the IR and RGB features respectively, the corresponding operations are defined as:
\begin{align}
M'_{T_{i}} &= \text{Conv}(f_{\mathrm{gus}}(M_{T_{i}})) \label{eq:fourier1} \\[-0.3em]
M'_{R_{i}} &= f_{\mathrm{fourier}}(M_{R_{i}}) \label{eq:fourier2}
\end{align}

\noindent
where \( M'_{T_{i}} \in \mathbb{R}^{B \times H \times W \times C_{\mathrm{out}}} \) and \( M'_{R_{i}} \in \mathbb{R}^{B \times H \times W \times C_{\mathrm{out}}} \) are the output features for the IR and RGB features, respectively. The function \( f_{\mathrm{fourier}}(\cdot) \) denotes the FourierKAN transformation applied to the RGB features, while \( f_{\mathrm{gus}}(\cdot) \) denotes the MSGKAN operation applied to the IR features. The operator \( \text{Conv}(\cdot) \) represents convolution operation.

\subsection{Spatial-frequency Feature Reconstruction (SFFR)}
\label{sec:SFFR}
Building upon the effectiveness of the aforementioned feature enhancement mechanisms, we propose a novel spatial-frequency feature reconstruction method, termed SFFR, as illustrated in Fig.~\ref{fig.3}. In this work, reconstruction refers to restoring and enhancing the potentially missing or underrepresented spatial and frequency information in RGB and IR features, thereby reconstructing a more complementary and discriminative representation. SFFR comprises two complementary sub-modules dedicated to intra-modal multi-scale feature learning and cross-modal feature enhancement, respectively. Finally, a dynamic weighting mechanism adaptively fuses the outputs of both branches, yielding a more robust and discriminative feature representation.

On the one hand, we adopt the MSGKAN module (Section~\ref{sec:MSGKAN feature enhancement}) to perform nonlinear feature modeling. By integrating KAN's nonlinear modeling with multi-scale Gaussian basis functions and convolution, MSGKAN reconstructs local structures and modality-specific semantic information in RGB and IR features, restoring spatial details that may be partially lost or weakened. This improves fine-grained feature representation and multi-scale structural complementarity, yielding enhanced interpretability and robustness to scale variations caused by differing UAV flight altitudes. On the other hand, we introduce the FCEKAN module (Section~\ref{sec:FourierKAN Component Exchange}), which transforms features into the frequency domain and selectively exchanges components across modalities within specific spectral bands. This process enables explicit cross-modal feature reconstruction in the frequency domain, effectively recovering complementary global patterns and promoting alignment between RGB and IR features.

By integrating the nonlinear modeling capabilities of KAN, the local feature extraction strengths of convolution, and the global frequency modeling capacity of Fourier decomposition, SFFR establishes a collaborative fusion framework that preserves modality-specific characteristics while effectively aggregating complementary information, thereby substantially enhancing the expressiveness of the fused representation. The formula is defined in (Eqs.~\eqref{eq:weighted_fused_R} and \eqref{eq:weighted_fused_T}):
\begin{align}
M_{R_{i}}' 
&= \alpha \cdot \text{Conv}\big(f_{\mathrm{gus}}(M_{R_{i}})\big) + \beta \cdot f_{\mathrm{Cross}}(M_{R_{i}}, M_{T_{i}})
\label{eq:weighted_fused_R} \\
M_{T_{i}}' 
&= \alpha \cdot \text{Conv}\big(f_{\mathrm{gus}}(M_{T_{i}})\big) + \beta \cdot f_{\mathrm{Cross}}(M_{T_{i}}, M_{R_{i}})
\label{eq:weighted_fused_T}
\end{align}

\noindent
where \(  M_{R_{i}} \) and \(  M_{T_{i}} \) denote the input features, \(  M_{T_{i}} '\) and \(  M_{R_{i}}' \) denote the input and output features. The weights \( \alpha \) and \( \beta \) are learnable parameters that dynamically adjust the contribution of each branch during training. The functions \( f_{\mathrm{gus}}(\cdot) \) and \( f_{\mathrm{Cross}}(\cdot) \) represent the operations of the MSGKAN and FCEKAN modules, respectively. The \( \text{Conv}(\cdot) \) denotes a standard convolution operation.

\subsection{Loss Function}
The traditional binary cross-entropy objective loss assigns more fixed weights to positive and negative samples when dealing with object confidence, which leads to low confidence positive samples being easily ignored, thus affecting the model's recall for object detection. To solve this problem, we introduce Varifocal Loss (VFLoss), which is able to adaptively adjust the focus on low-confidence positive samples by dynamically weighting the model so that it pays more attention to those positive samples with lower confidence. Its formula is as follows:\\[-0.5em]
\begin{equation}
{VFL}(p,q)=
\begin{cases}
-q (q \log(p) + (1-q)\log(1-p)) & q > 0 \\
- \alpha p^{\gamma} \log(1-p) & q = 0
\end{cases}
\label{vfl}
\end{equation}

\noindent
where $p$ is the predicted Instance-Aware Categorical Score (IACS) and $q$ is the object score. The IACS scalar is extracted from the categorical score vector: for each predicted bounding box, the ground truth category position contains its IoU with the ground truth bounding box, while all other positions are zero.

\section{Experiments}
\label{sec4}
\subsection{Datasets and Evaluation Metrics}
\textbf{SeaDroneSee Dataset: }SeaDroneSee~\cite{varga_seadronessee_2021} is a high-quality dataset designed for maritime UAV object detection and tracking tasks, covering a rich set of RGB-IR image pairs. Designed to support identification of critical maritime targets, it contains 246 training pairs and 61 validation pairs across seven categories: Swimmer, Floater, Boat, Human, Westperson, Lifejacket, and Ignored.

\textbf{DroneVehicle Dataset: }DroneVehicle~\cite{sun_drone-based_2021} is a large-scale RGB-IR vehicle detection dataset based on drones, containing 28439 pairs of RGB-IR images. It encompasses diverse urban scenarios (e.g., roads, parking lots) across both daytime and nighttime conditions. The dataset is partitioned into 17990 training pairs, 1469 validation pairs, and 8980 test pairs, covering five vehicle categories: Car, Truck, Bus, Van, and Freight\_car.

\textbf{DVTOD Dataset: }DVTOD~\cite{song_misaligned_2024} is the first UAV-captured misaligned RGB-IR object detection dataset based on a total of 2179 image pairs, containing data from varing times of the day and seasonal conditions. The dataset is partitioned into 1606 training pairs and 573 test pairs, covering three object categories: Car, Person, and Bicycle.

\textbf{Evaluation Metrics: }Average Precision (AP) represents the area under the precision-recall curve for a single category at specific IoU thresholds. Mean Average Precision (mAP) is computed as the mean of AP values across all categories. Common IOU thresholds include 0.5, 0.75, 0.5-0.95. mAP50 denotes the accuracy when the IOU threshold is 0.5, while mAP denotes a more comprehensive and stricter evaluation that averages AP across multiple thresholds from 0.5-0.95.

\subsection{Implementation Details}
The proposed method is implemented using PyTorch 2.0.1 framework on an Ubuntu 20.04 server equipped with an NVIDIA RTX-3080TI GPU and an CPU Intel i9-12900HX. The SGD optimizer is employed with the initial learning rate of 0.01 and with an momentum of 0.937. Additionally, the weight decay factor is set to 0.0005, and the learning rate follows a Cos annealing schedule. Training images are resized to $640 \times 640$ pixels, while test images are set to $640 \times 512$. Owing to its maturity, stability, extensive documentation, and broad adoption in multimodal detection tasks, YOLOv5 serves as a reliable foundation. Its lightweight design and fast inference make it especially suitable for real-time UAV applications.  Therefore, we adopt YOLOv5 with a two-branch design as the baseline framework and employ feature summation as the fusion strategy to ensure fair and consistent comparison with existing studies.

\subsection{Performance Comparison}
\textbf{SeaDroneSee Dataset: }Table~\ref{tab:sds} presents the results of our comparative experiments on the SeaDroneSee dataset. Although existing methods on this dataset are limited, our proposed approach outperforms them and achieves state-of-the-art performance in the key metric of mAP50. Compared to the baseline method, our method achieves a notable improvement of 4.9\%, 1.5\% and 1.1\% in terms of mAP50, mAP75 and mAP metrics. It is worth noting that although MCAFNet achieves a higher mAP75 of 33.3\%, our method surpasses it in both mAP50 and overall mAP, demonstrating stronger robustness across thresholds. Moreover, we can observe that ICAFusion performs relatively poorly on Swimmer, mainly due to their small size, weak features, and higher susceptibility to background interference, while achieving its best results on Floater with clearer structures and more distinct cross‑modal characteristics.
\begin{table}[H]
\centering
\caption{Comparison on the SeaDroneSee Dataset. Bold Numbers Indicate the Best Performance}
\label{tab:sds}
\scalebox{0.85}{
\begin{tabular}{lccc ccc cc}
\toprule
\multirow{2}{*}{Methods} & \multicolumn{3}{c}{AP50} & mAP50 & mAP75 & mAP & Params(M) & FLOPs(G) \\
\cmidrule(ll){2-4}
 & Swimmer & Floater & Boat &  &  &  &  &  \\
\midrule
DEYOLO~\cite{chen_deyolo_2025} & 13.5 & 35.7 & 98.0 & 49.2 & - & 27.8 & \textbf{78.3} & - \\
CFT~\cite{qingyun_cross-modality_2022} & 32.0 & 62.2 & 99.3 & 64.5 & - & 27.6&206.0&- \\
NIN Fusion~\cite{li_multispectral_2018} & \textbf{45.1} & 47.0 & \textbf{99.5} & 63.9 & - & 29.6 & - & - \\
ICAFusion~\cite{shen_icafusion_2024} & 14.7 &\textbf{69.5} & \textbf{99.5} & 61.2 & 29.1 & 30.5 & 120.2 & 100.6 \\
MCAFNet~\cite{zheng_mcafnet_2025}& 13.9 &63.2&99.3& 58.7 & \textbf{33.3}& 32.1& 243.1 &133.7\\
Baseline & 20.7 & 63.1 & \textbf{99.5} & 61.1 & 29.2 & 31.4 & 110.6 & \textbf{94.8}\\
Ours & 33.7 & 64.9 & \textbf{99.5} & \textbf{66.0} & 30.7 & \textbf{32.5} & 167.1 & 104.8\\
\bottomrule
\end{tabular}
}
\end{table}
\textbf{DroneVehicle Dataset: }Table~\ref{tab:DroneVehicle}  presents the comparative results on the DroneVehicle dataset. Our method attains a notable mAP50 of 84.4\%, demonstrating its superior detection performance. In terms of category-wise AP50, the proposed approach achieves 97.9\%, 96.6\%, 82.6\%, 74.0\%, and 71.1\% for the Car, Bus, Truck, Freight\_car, and Van categories, respectively. These results further substantiate the robustness capability of our model across diverse object classes.
\begin{table}[H]
\centering
\caption{
Comparison on the DroneVehicle Dataset}
\label{tab:DroneVehicle}
\scalebox{0.85}{
\begin{tabular}{lccccc ccc}
\toprule
\multirow{2}{*}{Methods} & \multicolumn{5}{c}{AP50} & mAP50 & Params(M) &FLOPs(G) \\
\cmidrule(ll){2-6}
  & Car & Truck & Bus & Van & Freight\_car &   &  &  \\
\midrule
MCMF~\cite{tian_mask-guided_2025} & 90.6 & 70.9 & 90.6 & 65.3 & 65.4 & 74.7 & 140.2 & - \\
C$^2$Former~\cite{yuan_mathbfc2former_2024} & 90.2 & 68.3 & 89.8 & 58.5 & 64.4 & 74.2 & 100.8 & 89.9 \\
CCLDet~\cite{shang_ccldet_2025} & 97.7 & 75.4 & 95.7 & 64.7 & 68.8 & 79.4 & 82.3 & 293.3 \\
ADMPF~\cite{liu_aerial_2025} & 97.9 & 82.2 & 95.9 & 59.5 & 69.3 & 82.0 & \textbf{31.8 }& 68.8 \\
ICAFusion~\cite{shen_icafusion_2024} & 97.8 &\textbf{84.2} &  96.2 & 69.6 & 71.9 & 84.1 & 120.2 & 100.6 \\
CMADet~\cite{song_misaligned_2024} & \textbf{98.2} & 70.4 & 78.3 &  \textbf{96.8} & 66.4 & 82.0 & 33.3 & \textbf{16.9} \\
GANNET~\cite{zheng_gaanet_2025} & 90.3 & 80.5 & 89.8 & 65.5 & 68.8 & 79.0 & 79.0 & 194.9 \\
MCAFNet~\cite{zheng_mcafnet_2025}& 98.1 &  82.2&\textbf{96.6}& 69.9 &71.3 & 83.6 & 243.1 &133.7\\
Baseline & 98.0 & 81.1 & 96.3 & 66.4 & 70.2 & 82.4 & 110.6 & 94.8 \\
Ours & 97.9 & 82.6 & \textbf{96.6} &71.1 & \textbf{74.0} & \textbf{84.4} & 167.1 &  104.8 \\
\bottomrule
\end{tabular}
}
\end{table}
\textbf{DVTOD Dataset: }Table~\ref{tab:DVTOD} shows the performance of our method compared to existing approaches on the DVTOD dataset. As shown, our approach consistently outperforms all competitors, achieving 88.0\%, 55.9\% and 51.9\% precision in terms of mAP50, mAP75 and mAP respectively. These results validate the superiority of our model in handling complex multimodal detection scenarios.
\begin{table}[H]
\centering
\caption{
Comparison on the DVTOD Dataset}
\label{tab:DVTOD}
\scalebox{0.85}{
\begin{tabular}{lccc ccc cc}
\toprule
\multirow{2}{*}{Methods} & \multicolumn{3}{c}{AP50} & mAP50 & mAP75 & mAP & Params(M) & FLOPs(G)\\
\cmidrule(ll){2-4}
 & Car & Bicycle & Person  &  &  &  &  &  \\
\midrule
YOLOv5+Add~\cite{song_misaligned_2024} & 88.8 & 74.3 & 74.6 & 79.2 & - & - & - & - \\
CMX~\cite{zhang_cmx_2023} & - & - & - & 82.2 & - & 42.3 & 139.9 & 134.3 \\
ICAFusion~\cite{shen_icafusion_2024} & - & - & - & 87.1 & - & 50.1 & 120.2 & 100.6 \\
CMADet~\cite{song_misaligned_2024} & 90.3 & 81.6 & 83.1 & 85.0 & - & - & \textbf{33.3} & \textbf{16.9} \\
GANNET~\cite{zheng_gaanet_2025}& - & - & -  & \textbf{88.0} & - & 51.2 & 79.0 & 194.9 \\
MCAFNet~\cite{zheng_mcafnet_2025}& 90.1 & 80.2 & 80.9  & 83.7 &52.5  & 49.4& 243.1 &133.7\\
Baseline    & 91.6 & 84.6 & \textbf{86.6} & 87.6 & 51.9 & 49.8 & 110.6 & 94.8 \\
Ours &\textbf{93.1} & \textbf{86.4} & 84.3 & \textbf{88.0} & \textbf{55.9} & \textbf{51.9} & 167.1 & 104.8\\
\bottomrule
\end{tabular}
}
\end{table}

\subsection{Ablation Studies}
\subsubsection{Effects of Each Module}
We have conducted an ablation study on the SeaDroneSee dataset to evaluate the performance of three core components. As shown in Table~\ref{tab:ablation_main}, each module positively impacts overall performance. Specifically, the FCEKAN module alone achieves 63.5\%, 31.7\%, and 31.9\% in terms of mAP50, mAP75, and mAP, respectively, demonstrating its effectiveness in capturing discriminative features through frequency-domain modeling. When used independently, MSGKAN improves mAP50, mAP75, and mAP by 1.1\%, 2.2\%, and 0.8\%, respectively, demonstrating the strong nonlinear modeling capability of KAN in capturing heterogeneous features across different modalities. Although incorporating FCEKAN and MSGKAN increases the parameters and FLOPs to 167.1M and 104.8G, respectively, their combination yields notable performance gains, achieving 64.4\% in mAP50 and 31.5\% in mAP, thereby demonstrating their synergistic effect through spatial-frequency reconstruction. With the additional integration of VFLoss, the model attains its best results, with 66.0\%, 30.7\%, and 32.5\% in terms of mAP50, mAP75 and mAP. These findings indicate that the joint contribution of the three components enables more comprehensive spatial-frequency reconstruction and substantially enhances cross-modal consistency, leading to a significant improvement in overall detection performance.
\begin{table}[H]
\centering
\caption{Effects of Each Module}
\label{tab:ablation_main}
\scalebox{0.8}{
\begin{tabular}{lcccccccccccc}
\toprule
\multirow{2}{*}{Index} & \multirow{2}{*}{FCEKAN}  & \multirow{2}{*}{MSGKAN} & \multirow{2}{*}{VFLoss} & \multicolumn{3}{c}{AP50} & \multirow{2}{*}{mAP50} & \multirow{2}{*}{mAP75} & \multirow{2}{*}{mAP} & \multirow{2}{*}{Params(M)} & \multirow{2}{*}{FLOPs(G)} \\
\cmidrule(ll){5-7}
                      &                          &                      &                        & Swimmer & Floater & Boat  &                        &                        &                      & & \\
\midrule
1 &   &   &   & 20.7 & 63.1 & 99.5 & 61.1 & 29.1 & 31.4 &\textbf{110.6 }& \textbf{94.8} \\
2 & \checkmark &   &   & 21.5& 69.7& 99.5& 63.5  & \textbf{31.7} & 31.9 &132.7 & 94.9 \\
3 &   & \checkmark &   & 15.1 & \textbf{71.9} & 99.5 & 62.2 & 31.3 & 32.2 &143.6 & 104.4 \\ 
4 &   &   & \checkmark & 26.4 & 61.7 & 99.5 & 62.5 & 26.0 & 29.6 &\textbf{110.6} &94.9\\
5 & \checkmark   & \checkmark  &   &22.9 & 70.8 & 99.5 & 64.4 & 29.6 & 31.5 & 167.1 & 104.8 \\
6 & \checkmark & \checkmark & \checkmark & \textbf{33.7} & 64.9 & 99.5 & \textbf{66.0} & 30.7 & \textbf{32.5} & 167.1 & 104.8 \\
\bottomrule
\end{tabular}}
\end{table}

\subsubsection{Comparison of FourierKAN and FCEKAN}
We conduct an ablation study to evaluate the impact of the proposed Fourier-domain component exchange strategy on model performance, as shown in Table~\ref{three_structures_results}. The experimental results demonstrate that this strategy exhibits notable advantages over the original FourierKan in frequency-domain modeling, achieving performance improvements of 2.4\%, 2.5\%, and 0.5\% over the baseline model in terms of mAP50, mAP75, and mAP, respectively. In addition, compared with the baseline, the number of parameters in FourierKAN and FCEKAN increases from 110.6M to 132.7M, while the FLOPs remain nearly unchanged.
These findings validate the effectiveness and superiority of the proposed component exchange mechanism in enhancing cross-modal feature fusion.

\begin{table}[H]
\centering
\small
\caption{Comparison of FourierKAN and FCEKAN modules}
\scalebox{0.85}{
\begin{tabular}{lcccccccccccc}
\toprule
\multirow{2}{*}{Methods} & \multicolumn{3}{c}{AP50}  & \multirow{2}{*}{mAP50} & \multirow{2}{*}{mAP75} & \multirow{2}{*}{mAP} & \multirow{2}{*}{Params(M)} & \multirow{2}{*}{FLOPs(G)} \\ 
\cmidrule(ll){2-4}
                       & Swimmer & Floater & Boat  & &  & & & \\
\midrule
Baseline       & 20.7 & 63.1 & 99.5 & 61.1 & 29.2 & 31.4 & \textbf{110.6} &\textbf{94.8} \\
FourierKAN     & 20.4 & 69.1 & 99.5 & 63.0 & 29.6 & \textbf{32.2} &132.7 & 94.9 \\
FCEKAN         & \textbf{21.5} & \textbf{69.7} & 99.5 & \textbf{63.5} & \textbf{31.7} & 31.9 & 132.7 & 94.9\\
\bottomrule
\end{tabular}
}
\label{three_structures_results}
\end{table}
\subsubsection{Comparison of FCEKAN Component Exchange Strategy}
In this section, as shown in Table~\ref{table:three_structures_results}, we perform a comparative analysis of the placement of FCEKAN components to elucidate the distinct contributions of sin and cos in feature modeling. The experimental results indicate that the integration of Fourier components in the order of sin(IR) and cos(RGB) provides significant advantages. Specifically, the model achieves 63.5\%, 31.7\%, and 31.9\% in terms of mAP50, mAP75, and mAP. These findings demonstrate the effectiveness of Fourier component configuration in enhancing cross-modal complementary feature representation.
\begin{table}[H]
\centering
\caption{Component exchange strategy in FCEKAN}
\scalebox{1}{
\begin{tabular}{lccccccc}
\toprule
{Methods} & \multicolumn{3}{c}{AP50}  & \multirow{2}{*}{mAP50} & \multirow{2}{*}{mAP75} & \multirow{2}{*}{mAP}   & \multirow{2}{*}{FLOPs(G)} \\ \cmidrule(ll){2-4}
                  & Swimmer & Floater & Boat & & &  & \\
\midrule
Baseline          &20.7~&63.1~&99.5~&61.1~&29.2~&31.4~&\textbf{94.8}\\
sin(RGB), cos(IR) &\textbf{23.1}~&65.9~&99.5~&62.8~&29.0~& 31.8~&94.9\\
sin(IR), cos(RGB) &21.5~&\textbf{69.7}~&99.5~&\textbf{63.5}~&\textbf{31.7}~&\textbf{31.9}~&94.9\\
\bottomrule
\end{tabular}}
\label{table:three_structures_results}
\end{table}

\subsubsection{Comparison of Scale Parameter}

In the Gaussian radial basis function, the scale parameter \( h \) determines the receptive field of the feature response. To systematically evaluate the impact of different scale combinations on model performance, we have designed and compared three sets of scale factors, which are detailed in Table~\ref{scale-param}. Experimental results in Table~\ref{scale-param} demonstrate that the \([1/7,\,3/7,\,5/7]\) setting, despite a minor increase in FLOPs to 104.4G, yields the best performance with 66.0\% mAP50, 30.7\% mAP75, and 32.5\% mAP. This configuration balances the scale intervals and the granularity of feature characterization, effectively enhancing the model’s capacity to capture scale variations induced by different UAV flight altitudes.

\begin{table}[H]
\centering
\footnotesize
\caption{Different Scale Parameter $h$}
\scalebox{1}{
\begin{tabular}{lcccccccc}
\toprule
\multirow{2}{*}{Scale (h)} & \multicolumn{3}{c}{AP50}  & \multirow{2}{*}{mAP50} & \multirow{2}{*}{mAP75} & \multirow{2}{*}{mAP}  & \multirow{2}{*}{FLOPs(G)} \\ \cmidrule(ll){2-4}
                       & Swimmer & Floater & Boat  & &  & & & \\
\midrule
$[1/3,2/3]$       & 21.4 & \textbf{68.6} & 99.5 & 63.2 & 30.0 & 30.0 &104.4 \\
$[1/5,3/5,5/5]$ &\textbf{35.6 }& 61.2& 99.5 & 65.0 & \textbf{32.8} & 32.3  & 104.4 \\
$[1/7,3/7,5/7]$ &33.7 & 64.9 & 99.5 & \textbf{66.0} & 30.7 & \textbf{32.5}  & 104.4 \\
$[1/9,3/9,5/9,7/9]$ &25.7 & 57.3 & 99.5 &60.8 & 20.1 & 27.0 & 104.4 \\
\bottomrule
\end{tabular}
}
\label{scale-param}
\end{table}

\subsubsection{Effects of Different Gridsize}
\begin{figure}[H]
    \centering
    \captionsetup{justification=justified}
    \includegraphics[width=1\textwidth]{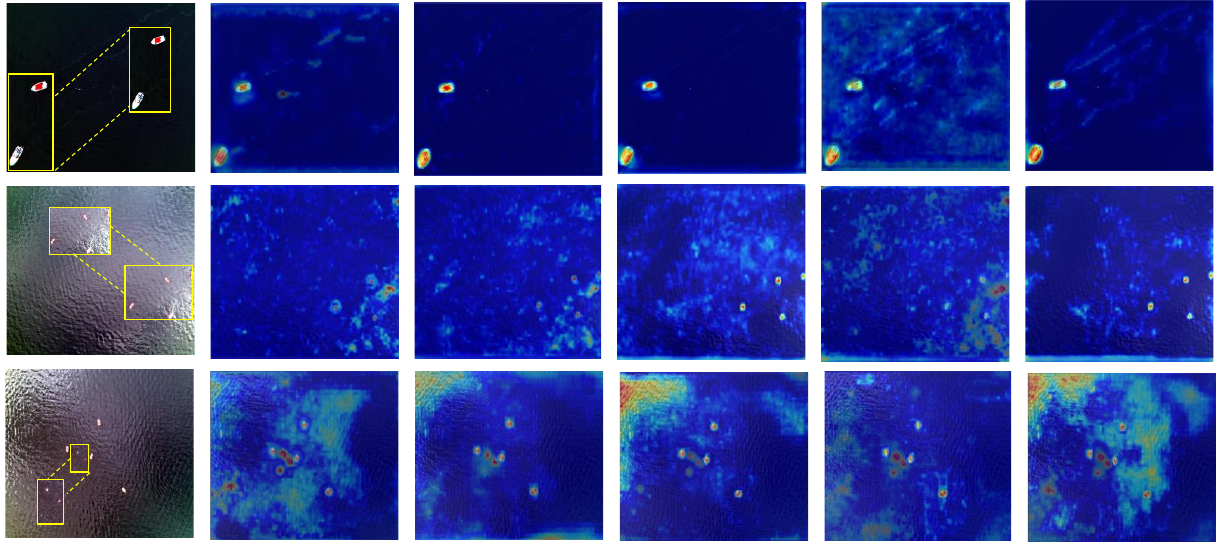}
    \caption{Visualization results with different gridsizes. The original image is presented in the leftmost panel, followed by the feature maps generated at gridsizes of 4, 8, 16, 32, and 64, arranged sequentially from left to right.}
    \label{figgridsize}
\end{figure}

To validate the impact of gridsize on model performance, we conducted experiments with varying gridsize. The experimental results, summarized in Table~\ref{tab:gridsearch}, indicate that the model achieves optimal performance at the gridsize of 8. As illustrated in Fig.~\ref{figgridsize}, visualization analysis reveals that gridsize plays a crucial role in balancing detail preservation in object detection. Specifically, an excessively large gridsize tends to capture redundant details and background noise, while an overly small gridsize risks missing critical semantic information. These observations underscore the importance of selecting an appropriate gridsize to ensure robust and accurate detection performance across diverse scenes.

\begin{table}[H]
\caption{Different Gridsize $g$}
\label{tab:gridsearch}
\centering\scalebox{1}{
\begin{tabular}{lccccc}
\toprule 
Gridsize (g) & mAP50 & mAP & Params(M) & FLOPs(G) & FPS(Hz) \\
\midrule
4   & 63.4 & 32.0 &\textbf{150.6}  & \textbf{100.1}  & \textbf{56.2} \\
8   & \textbf{66.0} & \textbf{32.5} & 167.1 &  104.8 & 33.3 \\
16  & 60.4 & 30.2 & 200.1 & 114.3  & 22.7 \\
32  & 62.2 & 31.0 & 266.2 & 133.4  & 14.1 \\
64  & 65.1 & 31.8 & 398.3 & 171.4  & 7.9  \\
\bottomrule
\end{tabular}}
\label{gridsize}
\end{table}

\subsubsection{Effects of Different Feature Modeling Methods}
Table~\ref{CNNtrans} presents an ablation study comparing the proposed KAN-based MSGKAN module with conventional CNN and Transformer blocks. Our method achieves the highest overall mAP of 31.5\% and Swimmer AP50 of 22.9\%, demonstrating the effectiveness of KAN-based modeling. Replacing MSGKAN with CNN or Transformer slightly reduces mAP to 31.4\% and 31.1\%, while CNN+CNN and Transformer+Transformer achieve 30.8\% and 31.0\%, indicating that standard blocks are less effective for cross-modal feature extraction. Transformer+Transformer attains the highest Floater AP50 of 74.3\%. These results confirm that MSGKAN improves detection performance across categories, particularly for challenging targets.

\begin{table}[H]
\centering
\small
\caption{Component Replacement with CNN and Transformer}
\label{CNNtrans}
\scalebox{1}{
\begin{tabular}{lcccccc}
\toprule
\multirow{2}{*}{Methods} & \multicolumn{3}{c}{AP50} & \multirow{2}{*}{mAP50} & \multirow{2}{*}{mAP75} & \multirow{2}{*}{mAP} \\ 
\cmidrule{2-4}
                       & Swimmer & Floater & Boat &  &  &  \\
\midrule
CNN+FCEKAN & 20.1 & 72.8 & 99.5 & 64.1 & 27.2 & 31.4 \\
Transformer+FCEKAN & 15.1 & 71.9 & 99.5 & 62.2 & \textbf{30.5} & 31.1 \\
CNN+CNN & 21.5 & 67.4 & 99.5 & 62.8 & \textbf{30.5} & 30.8 \\
Transformer+Transformer & 13.2 & \textbf{74.3} & 99.5 & 62.3 & 29.7 & 31.0 \\
Ours & \textbf{22.9} & 70.8 & 99.5 & \textbf{64.4} & 29.6 & \textbf{31.5} \\
\bottomrule
\end{tabular}
}
\end{table}

\subsubsection{Comparison of DETR and YOLO-based Methods}
Table~\ref{detr} presents a comparison between DETR and a YOLO-based detector for multispectral object detection. The results indicate that DETR achieves comparable AP50 and mAP, particularly for the Swimmer and Floater categories. Nevertheless, these performance levels are obtained at the expense of nearly ten times higher computational complexity, which limits its feasibility for UAV edge deployment. By contrast, the YOLO-based detector delivers competitive accuracy while significantly reducing parameter size and computational cost, thereby  offering a more balanced trade-off between accuracy and efficiency.

\begin{table}[H]
\centering
\small
\caption{Comparison of DETR and YOLO-based Methods}
\label{DETR}
\scalebox{1}{
\begin{tabular}{lccccccccc}
\toprule
\multirow{2}{*}{Methods} & \multicolumn{3}{c}{AP50}  & \multirow{2}{*}{mAP50} & \multirow{2}{*}{mAP75} & \multirow{2}{*}{mAP}   &\multirow{2}{*}{Params(M)} &\multirow{2}{*}{FLOPs(G)}\\ 
\cmidrule(ll){2-4}
                       & Swimmer & Floater & Boat  &  &  &  &  \\
\midrule
DETR        & \textbf{39.4} & \textbf{78.1} & 98.2 & \textbf{71.9} & \textbf{31.5} & \textbf{35.6} & 529.4 & 1059.0\\
YOLO-based & 33.7 & 64.9 & \textbf{99.5} & 66.0 & 30.7 & 32.5 & \textbf{167.1} & \textbf{104.8}\\
\bottomrule
\end{tabular}
}
\label{detr}
\end{table}

\subsection{Visual Interpretation}
\begin{figure}[H]
    \centering
    \captionsetup{justification=justified}
    \includegraphics[width=1\textwidth]{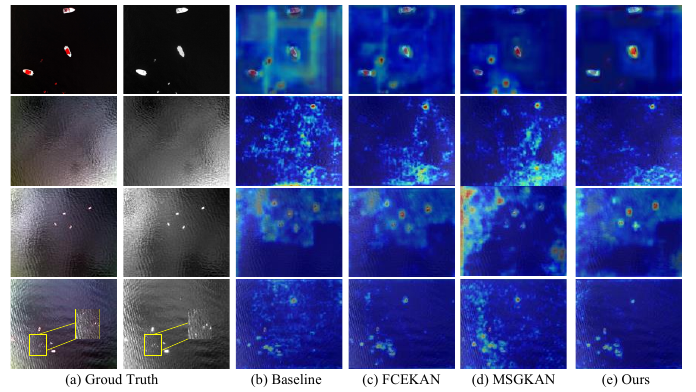}
    \caption{Visualization results of different components. From left to right are: ground truth in RGB and IR images, heatmaps of Baseline, FCEKAN, MSGKAN and our proposed method. Zoom in for more details.}
    \label{fig.5}
\end{figure}

\begin{figure}[H]
    \centering
\captionsetup{justification=justified} 
     \setlength{\abovecaptionskip}{1.9pt}
\includegraphics[width=1\textwidth]{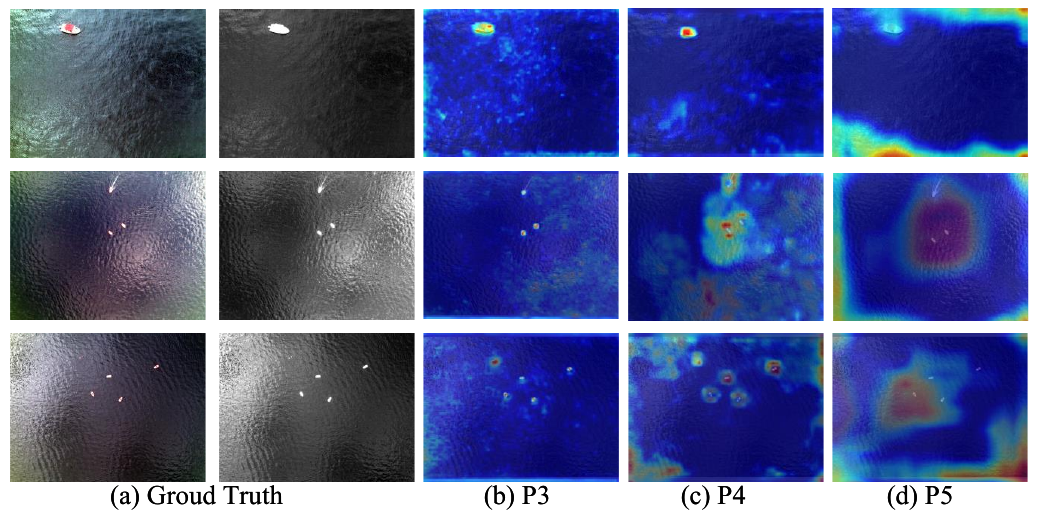}
    \caption{Visualization of the KANFusion at different feature stages. From left to right: ground truth in RGB and IR images, the P3-layer,the P4-layer, and the P5-layer outputs. }
    \label{KANFUSION}
\end{figure}

\begin{figure}[H]
    \centering
     \captionsetup{justification=justified} 
    \includegraphics[width=1\textwidth]{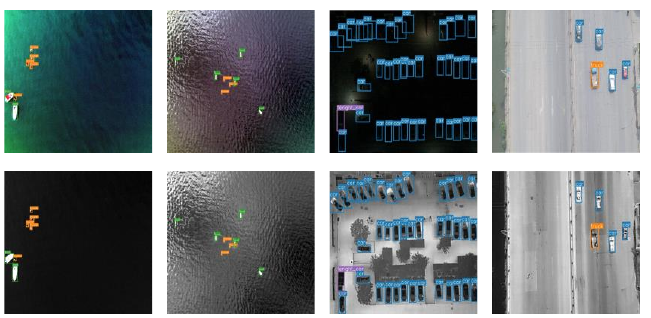}
    \caption{Sample detection results. From top to bottom, the rows correspond to RGB and IR modalities, and from left to right, the columns correspond to the SeaDroneSee and DroneVehicle datasets.}
    \label{fig.6}
\end{figure}

As shown in Fig.~\ref{fig.5} (a-e), the heatmaps are generated by the baseline, FCEKAN, MSGKAN, and our full method. The baseline suffers from scattered activations and false positives in complex backgrounds, particularly with water ripples or uneven lighting. In contrast, our method yields concentrated responses with clearer object boundaries, demonstrating superior localization and global perception. Furthermore, Fig.~\ref{KANFUSION} visualizes KANFusion at the P3, P4, and P5 layers: the P3-layer preserves fine details but introduces noise; the P4-layer balances contour clarity and noise suppression; and the P5-layer enhances semantic consistency, benefiting global perception. These results confirm that KANFusion achieves complementary modeling across feature stages, improving cross-modal fusion. As shown in Fig.~\ref{fig.6}, to provide a more intuitive evaluation of the detection results, we provide visualizations of some examples from the Seadronesee and DroneVehicle datasets.


\subsection{Limitations}
The proposed cross-modal and multi-scale feature fusion module is carefully designed to enable robust nonlinear representation learning and facilitate effective integration of RGB and IR modality features. Nonetheless, certain limitations remain. While the shallow architecture mitigates issues related to deep stacking, the presence of multiple branches—particularly the KAN mappings—results in considerable computational and memory overhead at high input resolutions. Future work focuses on investigating more efficient frequency band partitioning mechanisms and advanced cross-modal geometric alignment strategies to further improve the model’s generalization ability and applicability in multi-modal environments.

\section{Conclusion}
\label{sec5}
In this paper, we propose a novel cross-modal and multi-scale feature fusion framework for aerial multispectral object detection, aiming to address the inefficiencies in integrating heterogeneous modality information. The proposed framework innovatively combines spatial- and frequency-domain fusion strategies, leveraging the powerful nonlinear modeling capability of the KAN architecture and the distinctive advantages of Fourier transform in frequency analysis. Extensive experiments conducted on the SeaDroneSee, DroneVehicle, and DVTOD datasets demonstrate the superior performance and effectiveness of our approach compared to existing methods. Unlike conventional neural network architectures, whose fixed kernels or attention mechanisms constrain their ability to capture fine spatial–frequency variations, the KAN-based design offers enhanced flexibility in modeling complex nonlinear interactions, thereby making it a promising choice for enhancing multispectral object detection. In future work, we plan to further explore model lightweighting and efficient inference strategies to reduce computational and memory overhead while maintaining detection accuracy, thereby enhancing the framework’s practicality for real-world deployment.

\section*{Acknowledgments}
This work was supported in part by National Natural Science Foundation of China under Grant No. 61903164 and in part by Natural Science Foundation of Jiangsu Province in China under Grants
BK20191427 and the Key R\&D Program of Zhejiang Province (2024C04056(CSJ)).

\ifCLASSOPTIONcaptionsoff
  \newpage
\fi






% \end{thebibliography}
\bibliographystyle{IEEEtran}
\bibliography{ref}



%\vfill


\end{document}